\documentclass[runningheads]{llncs}
\usepackage{graphicx}
\usepackage[symbol]{footmisc}

\usepackage{tikz}
\usepackage{comment}
\usepackage{amsmath,amssymb} 
\usepackage{color}
\usepackage{multirow}
\usepackage{makecell}
\usepackage{bbding}
\usepackage[colorlinks]{hyperref}

\usepackage[accsupp]{axessibility}  

\usepackage[width=122mm,left=12mm,paperwidth=146mm,height=193mm,top=12mm,paperheight=217mm]{geometry}

\begin{document}
\pagestyle{headings}
\mainmatter
\def\ECCVSubNumber{669}  

\title{SeqTR: A Simple yet Universal Network for Visual Grounding} 

\titlerunning{SeqTR: A Simple yet Universal Network for Visual Grounding}
\author{
Chaoyang Zhu$^{1}$, 
Yiyi Zhou$^{1}$, 
Yunhang Shen$^{3}$, 
Gen Luo$^{1}$, 
Xingjia Pan$^{3}$, \\ 
Mingbao Lin$^{3}$, 
Chao Chen$^{3}$, 
Liujuan Cao$^{1*}$, 
Xiaoshuai Sun$^{1,4}$, 
Rongrong Ji$^{1,2,4}$
}
\authorrunning{C. Zhu et al.}
\institute{
$^{1}$MAC Lab, Department of Artificial Intelligence, School of Informatics, Xiamen University.
~$^{2}$Institute of Energy Research, Jiangxi Academy of Sciences.\\
$^{3}$Tencent Youtu Lab.
~$^{4}$Institute of Artificial Intelligence, Xiamen University.\\
\email{cyzhu@stu.xmu.edu.cn}, 
\email{zhouyiyi@xmu.edu.cn}, 
\email{shenyunhang01@gmail.com}, \\
\email{luogen@stu.xmu.edu.cn}, 
\email{xjia.pan@gmail.com}, 
\email{linmb001@outlook.com}, \\
\email{aaronccchen@tencent.com}, 
\email{\{caoliujuan,xssun,rrji\}@xmu.edu.cn}}
\renewcommand{\footnotemark}{\fnsymbol{footnote}}
\footnotetext[1]{Corresponding author.}
\renewcommand{\thefootnote}{\arabic{footnote}}

\maketitle

\begin{abstract}
In this paper, we propose a simple yet universal network termed \emph{SeqTR} for visual grounding tasks, \emph{e.g.}, phrase localization, referring expression comprehension (REC) and segmentation (RES). The canonical paradigms for visual grounding often require substantial expertise in designing network architectures and loss functions, making them hard to generalize across tasks. To simplify and unify the modeling, we cast visual grounding as a point prediction problem conditioned on image and text inputs, where either the bounding box or binary mask is represented as a sequence of discrete coordinate tokens. Under this paradigm, visual grounding tasks are unified in our SeqTR network without task-specific branches or heads, \emph{e.g.}, the convolutional mask decoder for RES, which greatly reduces the complexity of multi-task modeling. In addition, SeqTR also shares the same optimization objective for all tasks with a simple \emph{cross-entropy} loss, further reducing the complexity of deploying hand-crafted loss functions. Experiments on five benchmark datasets demonstrate that the proposed SeqTR outperforms (or is on par with) the existing state-of-the-arts, proving that a simple yet universal approach for visual grounding is indeed feasible. Source code is available at \href{https://github.com/sean-zhuh/SeqTR}{https://github.com/sean-zhuh/SeqTR}.
\end{abstract}

\section{Introduction}
\label{sect:introduction}

Visual grounding~\cite{refcoco/+,refcocoggoogle,refcocogumd,referitgame,flickr30k} has emerged as a core problem in vision-language research, as both comprehensive intra-modality understanding and accurate one-to-one inter-modality correspondence establishment are required. According to the manner of grounding, it can be divided into two groups, \emph{i.e.}, \emph{phrase localization} or \emph{referring expression comprehension} (REC) at bounding box level~\cite{mattnet,cmatterase,nmtree,realgin,faoa,rccf,mcn,resc,iterativeshrinking,lbyl,transvg,trar,reftr}, and \emph{referring expression segmentation} (RES) at pixel level~\cite{mattnet,cmsa,step,brinet,lscm,cmpc+,mcn,efn,busnet,cgan,lts,vlt,reftr}.

To accomplish the accurate vision-language alignment, existing approaches often require substantial prior knowledge and expertise in designing network architectures and loss functions. For instance, MAttNet~\cite{mattnet} decomposes language expressions into \emph{subject}, \emph{location}, and \emph{relationship} phrases, and designs three corresponding attention modules to compute matching score individually. Despite being faster, one-stage models also require the complex language-guided multi-modal fusion and reasoning modules~\cite{laconv,lbyl,rccf,resc,mcn}, or sophisticated cross-modal alignment via various attention mechanisms~\cite{efn,cgan,cmsa,brinet,cmpc+,mcn,vlt}. Loss functions in existing methods are also complex and tailored to each individual grounding task, such as GIoU loss~\cite{giou}, set-based matching loss~\cite{detr}, focal loss~\cite{focal}, dice loss~\cite{dice}, and contrastive alignment loss~\cite{mdetr}. Under a multi-task setting, coefficients among different losses also need to be carefully tuned to accommodate different tasks~\cite{mcn,reftr}. Despite great progress, these highly customized approaches still suffer from the limited generalization ability.

\begin{figure*}[t]
\centering
\includegraphics[height=2.41cm, width=12.2cm]{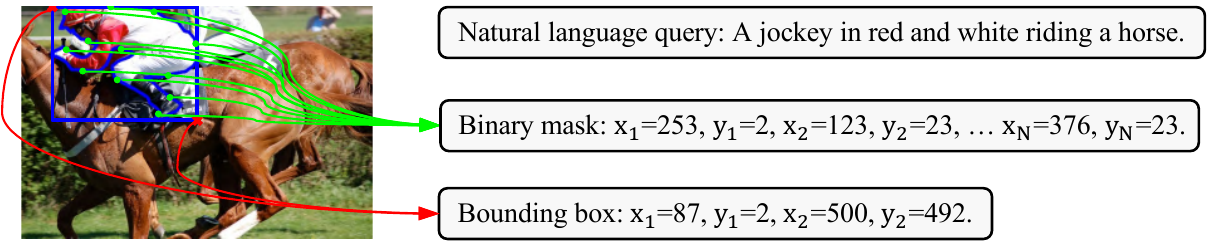}
\caption{Illustration of the serialization of grounding information. Our model directly generates the sequence of points representing the bounding box or binary mask.}
\label{fig:overview}
\end{figure*}

Recent endeavors~\cite{transvg,vlt,mdetr,reftr} in visual grounding shift to simplifying network architectures via Transformers~\cite{transformer}. Concretely, the multi-modal fusion and reasoning modules are replaced by a simple stack of transformer encoder layers~\cite{transvg,mdetr,vlt}. However, the loss function used in these transformer-based methods is still highly customized for each individual task~\cite{focal,dice,mdetr,giou,detr}. Moreover, these approaches still require task-specific branches or heads~\cite{mcn,reftr}, \emph{i.e.}, the bounding box regressor and convolutional mask decoder.

In this paper, we take a step forward in simplifying the modeling of visual grounding tasks via a simple yet universal network termed \emph{SeqTR}. Specifically, inspired by the recently proposed Pix2Seq~\cite{pix2seq}, we first reformulate visual grounding as a point prediction problem conditioned on image and text inputs, where the grounding information, \emph{e.g.}, the bounding box, is serialized into a sequence of discrete coordinate tokens. Under this paradigm, different grounding tasks can be universally accomplished in the proposed SeqTR with a standard transformer encoder-decoder architecture~\cite{transformer}. In SeqTR, the encoder serves to update the multi-modal feature representations, while the decoder directly predicts the discrete coordinate tokens of the grounding information in an auto-regressive manner. In terms of optimization, SeqTR only uses a simple \emph{cross-entropy} loss for all grounding tasks, requiring no further prior knowledge or expertise. Overall, the proposed SeqTR greatly reduces the difficulty and complexity of both architecture design and optimization for visual grounding. 

Notably, the proposed SeqTR is not just a simple multi-modal extension of Pix2Seq for the challenging open-ended visual grounding tasks. In addition to bridging the gap between object detection and visual grounding, we also apply the sequential modeling to RES via an innovative \emph{mask contour sampling} scheme. As shown in Fig.~\ref{fig:overview}, SeqTR transforms the pixel-wise binary mask into a sequence of $N$ points by performing clockwise sampling on the mask contour. In this case, RES, as a language-guided segmentation task, can be seamlessly integrated into the proposed SeqTR network without the additional convolutional mask decoder, demonstrating the high generalization ability of SeqTR across grounding tasks.

The proposed SeqTR achieves or is on par with the state-of-the-art performance on five benchmark datasets, \emph{i.e.}, RefCOCO~\cite{refcoco/+}, RefCOCO+~\cite{refcoco/+}, RefCOCOg~\cite{refcocoggoogle,refcocogumd}, ReferItGame~\cite{referitgame}, and Flickr30K Entities~\cite{flickrentities}. SeqTR also outperforms a set of large-scale BERT-style models~\cite{vilbert,vlbert,uniter,mdetr} with much less pre-training expenditure. Main contributions are summarized as follows:
\begin{itemize}
    \item We reformulate visual grounding tasks as a point prediction problem, and present a novel and general network, termed SeqTR, which unifies different grounding tasks in one model with the same \emph{cross-entropy} loss. 
    \item The proposed SeqTR is simple yet universal, and can be seamlessly extended to the referring expression segmentation task via an innovative \emph{mask contour sampling} scheme without network architecture modifications.
    \item We achieve or maintain on par with the state-of-the-art performance on five visual grounding benchmark datasets, and also outperform a set of large-scale pre-trained models with much less expenditure. 
\end{itemize}

\section{Related Work}
\label{sect:related work}

\subsection{Referring Expression Comprehension}
Early practitioners~\cite{cmn,vc,parallelattention,mattnet,cmatterase,dga,rvgtree,nmtree} tackle referring expression comprehension (REC) following a two-stage pipeline, where region proposals~\cite{fasterrcnn} are first extracted then ranked according to their similarity scores with the language query. Another line of work~\cite{realgin,faoa,rccf,mcn,resc,iterativeshrinking,lbyl,transvg,trar}, being simpler and faster, advocates one-stage pipeline based on dense anchors~\cite{fasterrcnn}. RealGIN~\cite{realgin} proposes adaptive feature selection and global attentive reasoning unit to handle the diversity and complexity of language expressions. ReSC~\cite{resc} recursively constructs sub-queries to predict the parameters of the normalization layers in the visual encoder, which is used to scale and shift visual features. LBYL~\cite{lbyl} designs landmark feature convolution to encode the contextual information. Recent works~\cite{transvg,trar,mdetr,vlt,reftr} resort to Transformer-like structure~\cite{transformer} to perform multi-modal fusion. MDETR~\cite{mdetr} further demonstrates that Transformer is efficient when pre-trained on a large corpus of data. Compared with existing approaches, our work is simple in both the architecture and loss function, which has little requirement of task priors and expert engineering.

\subsection{Referring Expression Segmentation}
Compared to REC, referring expression segmentation (RES) grounds language query at a fine-granularity \emph{i.e.}, the precise pixel-wise binary mask. Typical solutions are to design various attention mechanisms to perform cross-modal alignment~\cite{cgan,mcn,vlt,cmsa,efn,step,brinet,cmpc,cmpc+,lscm}. EFN~\cite{efn} transforms the visual encoder into a multi-modal feature extractor with asymmetric co-attention, which fuses multi-modal information at the feature learning stage. CGAN~\cite{cgan} performs cascaded attention reasoning with instance-level attention loss to supervise attention modeling at each stage. LTS~\cite{lts} first performs relavance filtering to locate the referent, and uses this visual object prior to perform dilated convolution for the final segmentation mask. VLT~\cite{vlt} produces a set of queries representing different understandings of the language expression and proposes a query balance module to focus on the most reasonable and suitable query, which is then used to decode the mask via a mask decoder. In this work, we are the first to regard RES as a point prediction problem, thus the proposed SeqTR can be seamlessly extended to RES without any network architecture modifications.

\subsection{Multi-task Visual Grounding}
Multi-task visual grounding aims to jointly address REC and RES. Prior art MCN~\cite{mcn} constrains the REC and RES branches to attend to the same region by applying consistent energy maximization. In this way, REC can help RES better localize the referent, and RES can help REC achieve superior cross-modal alignment. RefTR~\cite{reftr} tackles multi-task visual grounding by sharing the same transformer architecture, but it requires an additional convolutional mask decoder for RES. In contrast, the proposed SeqTR is universal across different grounding tasks without additional branch or head. Under the point prediction paradigm, SeqTR can segment the referent without the aid from REC branch.

\section{Method}
\label{sect:method}

In this section, we introduce our simple yet universal SeqTR network for visual grounding, of which structure is depicted in Fig.~\ref{fig:architecture}. The objective function is detailed in Sec.~\ref{sect:point prediction}. Sequence construction from grounding information is elaborated in Sec.~\ref{sect:sequence construction}. The architecture and inference are presented in Sec.~\ref{sect:architecture}.

\subsection{Problem Definition}
\label{sect:point prediction}

\begin{figure*}[t]
\centering
\includegraphics[height=3.3cm, width=12.2cm]{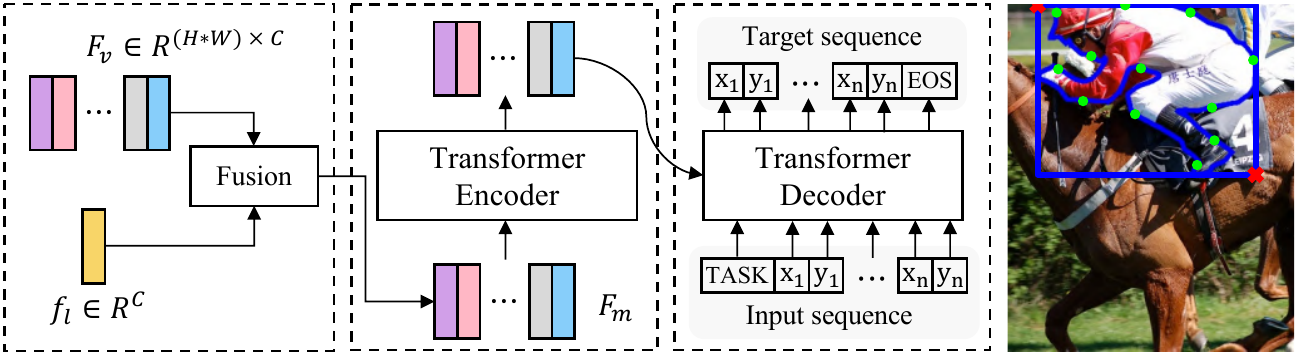}
\caption{Overview of the proposed SeqTR network, of which all components, \emph{i.e.}, multi-modal fusion, cross-modal interaction, and loss function, are standard operations and shared across grounding tasks.}
\label{fig:architecture}
\end{figure*}

Unlike existing visual grounding models~\cite{mcn,transvg,efn,lts,vlt}, SeqTR aims to predict the discrete coordinate tokens of the grounding information, \emph{e.g.}, the bounding box or binary mask. To this end, we define the optimization objective under the point prediction paradigm as:
\begin{equation}
\label{eq:loss}
    \mathcal{L} = -\sum_{i=1}^{2N}w_{i}\log P(T_{i}|F_m, S_{1:i-1}),
\end{equation}
where $S$ and $T$ are the input and target sequences for decoder as shown in Fig.~\ref{fig:architecture}. $F_m \in R^{(H*W) \times C}$ is the multi-modal features detailed in Sec.~\ref{sect:architecture}. A per-token weight $w_i$ is used to scale the loss. Note that the input sequence $S_{1:i-1}$ only contains the preceding coordinate tokens when predicting the $i$-th one. It can be implemented by putting a causal mask~\cite{gpt2} on attention weights to only attend to previous coordinate tokens. \par
We construct the input sequence by prepending a $\left[\text{TASK}\right]$ token before the sequence of points $\{x_i, y_i\}_{i=1}^{N}$, and the target sequence is the one appended with an $\left[\text{EOS}\right]$ token. These two special tokens indicate the start or end of the sequence, which are learnable embeddings. $\left[\text{TASK}\right]$ token also indicates which grounding task the model performs on. To achieve multi-task visual grounding, we can equip each task with the corresponding $\left[\text{TASK}\right]$ token randomly initialized with different parameters, showing great simplicity and generalization ability. \par
Under our point prediction reformulation, the simple \emph{cross-entropy} loss conditioned on multi-modal features and preceding discrete coordinate tokens can be directly shared across tasks, avoiding the complex deployment of hand-crafted loss functions and loss coefficient tuning~\cite{dice,focal,detr,mdetr,giou}.

\subsection{Sequence Construction from Grounding Information}
\label{sect:sequence construction}

A key design in SeqTR is to serialize and quantize the grounding information, \emph{e.g.}, the bounding box or binary mask, into a sequence of discrete coordinate tokens, which enables different grounding tasks to be universally addressed in one network architecture with the same objective. \par
We first review the serialization and quantization of the bounding box introduced in Pix2Seq~\cite{pix2seq}. Given a sequence of floating points $\{\tilde{x_i}, \tilde{y_i}\}_{i=1}^{N}$ representing the top-left and bottom-right corner points of the bounding box ($N$ is 2), these floating coordinates are quantized into integer bins by
\begin{equation}
\label{eq:quantization}
    x_i = \text{round}(\frac{\tilde{x_i}}{w} * M),\qquad
    y_i = \text{round}(\frac{\tilde{y_i}}{h} * M),
\end{equation}
where each coordinate is normalized by image width $w$ and height $h$, and $M$ is the number of quantization bins. We refer readers to Pix2Seq~\cite{pix2seq} for more discretization details. In practice, we construct a shared embedding vocabulary $E \in R^{M \times C}$ for both $x$-axis and $y$-axis. \par
While bounding boxes can be naturally determined by two of its corner points and serialized into a sequence as in Eq.~\ref{eq:quantization}, binary masks can not. A binary mask consists of infinite points, of which both quantities and positions impact the details of the mask significantly, thus the above serialization and quantization for bounding boxes is not directly applicable to binary masks. \par
\begin{figure*}[t]
\centering
\includegraphics[height=2.01cm, width=12.05cm]{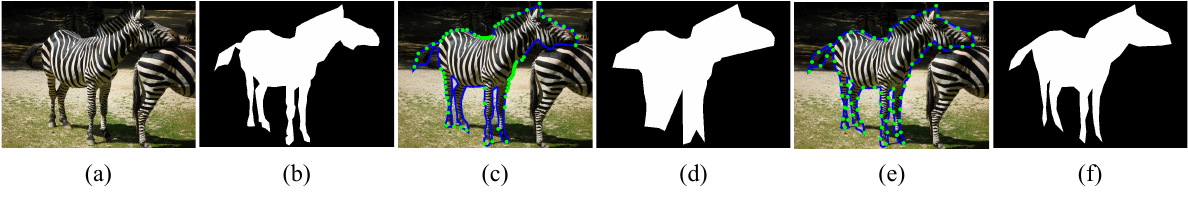}
\caption{Visualization of different sampling strategies. (a-b) are the original image and ground-truth. (c-d) are the sampled points and reassembled mask of center-based sampling, respectively, while (e-f) are the ones of uniform sampling.}
\label{fig:sampling}
\end{figure*}

To address this issue, we propose an innovative \emph{mask contour sampling} scheme for the sequence construction from binary masks. As shown in Fig.~\ref{fig:sampling}, we sample $N$ points clockwise from the consecutive mask contour of the referred object, then, the sequence of sampled points can be quantized via Eq.~\ref{eq:quantization}. Following sampling strategies are experimented:
\begin{itemize}
    \item\textbf{Center-based sampling}. Starting from the mass center of the binary mask, $N$ rays are emitted with the same angle interval. The intersection points between these rays and the mask contour are clockwise sampled.
    \item\textbf{Uniform sampling}. We uniformly sample $N$ points clockwise on top of the mask contour, which is much simpler compared to the first strategy.
\end{itemize}
Compared to the center-based sampling, uniform sampling distributes the sampled points along the mask contour more evenly, and can better represent the irregular mask especially when the outline between two adjacent sampled points is tortuous. As shown in Fig.~\ref{fig:sampling}, center-based sampling loses the fine details of the zebra legs, while uniform sampling preserves the mask contour more precisely. \par
In practice, the proposed sampling scheme slightly restricts the performance upper-bound of RES, \emph{e.g.}, uniformly sampling 36 points from ground-truth masks will achieve 95.63 mIoU on RefCOCO \emph{validation} set. Considering current state-of-the-art performance, such a defect is still acceptable. Besides, even if we take as ground-truth the precise binary mask, the upper-bound still will not reach 100 mIoU since down-sampling operations are often necessary. \par
Both center-based and uniform sampling use deterministic (clockwise) ordering in the sequence of points for the binary mask, however, a binary mask is only determined by points’ positions instead of the ordering. Hence we randomly shuffle points’ order, which enables the model to learn which point to predict next. In Sec.~\ref{sect:ablationstudies}, we thoroughly study the proposed sampling scheme.

\subsection{Architecture}
\label{sect:architecture}

\noindent{\bf Language Encoder.} To demonstrate the efficacy of SeqTR, we do not opt for the pre-trained language encoders such as BERT~\cite{bert}, hereby the language encoder is a one layer bidirectional GRU~\cite{gru}. We concatenate both unidirectional hidden states $h_t=[\overrightarrow{h_t};\overleftarrow{h_t}]$ at each step $t$ to form word features $\{h_t\}_{t=1}^{T}$.

\noindent{\bf Visual Encoder.} The multi-scale features of the visual encoder are unidirectionally down-sampled from the finest to coarsest spatial resolution, and flattened to generate visual features $F_v \in R^{(H*W) \times C}$ as input to the fusion module. $H$ and $W$ are 32 times smaller of the original image size. In contrast to previous work, we only use the coarsest scale visual features instead of the finest ones for RES task~\cite{mcn,lts,efn,cgan}, as we do not predict the binary mask pixel-wisely, which reduces the memory footprint during training.

\noindent{\bf Fusion.} Different from Pix2Seq~\cite{pix2seq}, which only perceives the pixel inputs, we devise a simple yet efficient fusion module to align vision and language modalities. Given visual features $F_v$ and word features $\{h_t\}_{t=1}^T$, we first construct language feature $f_l \in R^{C}$ by \emph{max} pooling word features along the channel dimension. We use Hadamard product between $F_v$ and $f_l$ without the linear projection to produce the multi-modal features $F_m \in R^{(H*W) \times C}$ to transformer encoder:
\begin{equation}
\label{eq:fusion}
    F_{m,i} = \sigma(F_{v,i}) \odot \sigma(f_l),
\end{equation}
where $\sigma$ is tanh function. Note that we do not concatenate word features and visual features then use the transformer encoder to perform fusion as in~\cite{transvg,mdetr,vlt}, because that the complexity will quadratically increase.

\noindent{\bf Transformer and Predictor.} The standard transformer encoder updates the feature representations of multi-modal features $F_m$, while the decoder predicts the target sequence in an auto-regressive manner. The hidden dimension of transformer is set to 256, the expansion rate in feed forward network (FFN) is 4, and the number of encoder and decoder layers are 6 and 3, respectively. This results in the transformer being extremely compact. Since the transformer is permutation-invariant, the $F_m$ and the input sequence are added with \emph{sine} and \emph{learned} positional encoding~\cite{transformer}, respectively. To predict the coordinate tokens, an MLP with a final softmax function is used.

\noindent{\bf Inference.} During inference, coordinates are generated in an auto-regressive manner, each coordinate is the \emph{argmax}-ed index of the probabilities over the vocabulary $E$, and mapped back to the original image scale via the inversion of Eq.~\ref{eq:quantization}. We predict exactly 4 discrete coordinate tokens for REC, while leaving the decision of when to prediction to $\left[\text{EOS}\right]$ token for RES. The predicted sequence is assembled to form the bounding box or binary mask for evaluation.  

\section{Experiments}
\label{sect:experiments}

\subsection{Datasets}
\label{sect:datasets}

\noindent{\bf RefCOCO/RefCOCO+/RefCOCOg.}
    RefCOCO~\cite{refcoco/+} contains 142,210 referring expressions, 50,000 referred objects, and 19,994 images. Referring expressions in testA set mostly describe people, while the ones in testB set mainly describe objects except people. Similarly, RefCOCO+~\cite{refcoco/+} contains 141,564 expressions, 49,856 referred objects, and 19,992 images. Compared to RefCOCO, referring expressions of RefCOCO+ describe more about attributes of the referent, \emph{e.g.}, color, shape, digits, and avoid using words of absolute spatial location. RefCOCOg~\cite{refcocoggoogle,refcocogumd} has two types of partition strategy, \emph{i.e.}, the \emph{google} split~\cite{refcocoggoogle} and \emph{umd} split~\cite{refcocogumd}. Both splits have 95,010 referring expressions, 49,822 referred objects, and 25,799 images. We use the validation set as the test set following ~\cite{efn,busnet,lbyl,resc} for \emph{umd} split. The language length of RefCOCOg is 8.4 words on average while that of RefCOCO and RefCOCO+ are only 3.6 and 3.5 words.
    
\noindent{\bf ReferItGame}~\cite{referitgame} contains 120,072 referring expressions and 99,220 referents for 19,997 images collected from the SAIAPR-12~\cite{saiapr12} dataset. We use the cleaned berkeley split to partition the dataset, which consists of 54,127, 5,842, and 60,103 referring expressions in train, validation, and test set, respectively.

\noindent{\bf Flickr30K.} Language queries in Flickr30K Entities~\cite{flickrentities} are short region phrases instead of sentences which may contain multiple objects. It contains 31,783 images with 427K referred entities in train, validation, and test set.

\noindent{\bf Pre-training dataset.} Following~\cite{mdetr}, we merge region descriptions from Visual Genome (VG)~\cite{vg} dataset, annotations from RefCOCO~\cite{refcoco/+}, RefCOCO+~\cite{refcoco/+}, RefCOCOg~\cite{refcocoggoogle,refcocogumd}, and ReferItGame~\cite{referitgame} datasets, and Flickr entities~\cite{flickrentities}. This results in approximately 6.1M distinct language expressions and 174k images in train set, which are less than 200k images as in~\cite{mdetr}.

\setlength{\tabcolsep}{2.0pt}
\begin{table*}[t]
\scriptsize
\begin{center}
\caption{Comparison with the state-of-the-arts on the REC task. Visual encoders of models with \dag~is trained without excluding val/test images of the three datasets. RN101 refers to ResNet101~\cite{resnet} and DN53 denotes DarkNet53~\cite{yolov3}.}
\label{table:sotaonrec}
\begin{tabular}{lcccc|ccc|ccc|c}
\hline\noalign{\smallskip}
\multirow{2}{*}{Models} & \multirow{2}{*}{\makecell{Visual \\ Encoder}} & \multicolumn{3}{c}{RefCOCO} & \multicolumn{3}{c}{RefCOCO+} & \multicolumn{3}{c}{RefCOCOg} & Time \\
& & val & testA & testB & val & testA & testB & val-g & val-u & test-u & (ms) \\
\noalign{\smallskip}
\hline
\noalign{\smallskip}
\multicolumn{12}{l}{\textbf{Two-stage}} \\
\noalign{\smallskip}
\hline
\noalign{\smallskip}
CMN~\cite{cmn} & VGG16 & - & 71.03 & 65.77 & - & 54.32 & 47.76 & 57.47 & - & - & - \\
VC~\cite{vc} & VGG16 & - & 73.33 & 67.44 & - & 58.40 & 53.18 & 62.30 & - & - & - \\
ParalAttn~\cite{parallelattention} & VGG16 & - & 75.31 & 65.52 & - & 61.34 & 50.86 & 58.03 & - & - & - \\
MAttNet~\cite{mattnet} & RN101 & 76.40 & 80.43 & 69.28 & 64.93 & 70.26 & 56.00 & - & 66.58 & 67.27 & 320 \\
CM-Att-Erase~\cite{cmatterase} & RN101 & 78.35 & 83.14 & 71.32 & 68.09 & 73.65 & 58.03 & - & 67.99 & 68.67 & - \\
DGA~\cite{dga} & VGG16 & - & 78.42 & 65.53 & - & 69.07 & 51.99 & - & - & 63.28 & 341 \\
RvG-Tree~\cite{rvgtree} & RN101 & 75.06 & 78.61 & 69.85 & 63.51 & 67.45 & 56.66 & - & 66.95 & 66.51 & - \\
NMTree~\cite{nmtree} & RN101 & 76.41 & 81.21 & 70.09 & 66.46 & 72.02 & 57.52 & 64.62 & 65.87 & 66.44 & - \\
\noalign{\smallskip}
\hline
\noalign{\smallskip}
\multicolumn{12}{l}{\textbf{One-stage}} \\
\noalign{\smallskip}
\hline
\noalign{\smallskip}
RealGIN~\cite{realgin} & DN53 & 77.25 & 78.70 & 72.10 & 62.78 & 67.17 & 54.21 & - & 62.75 & 62.33 & 35 \\
$\text{FAOA}^{\dag}$~\cite{faoa} & DN53 & 71.15 & 74.88 & 66.32 & 56.86 & 61.89 & 49.46 & - & 59.44 & 58.90 & 39 \\
RCCF~\cite{rccf} & DLA34 & - & 81.06 & 71.85 & - & 70.35 & 56.32 & - & - & 65.73 & \textbf{25} \\
MCN~\cite{mcn} & DN53 & 80.08 & 82.29 & 74.98 & 67.16 & 72.86 & 57.31 & - & 66.46 & 66.01 & 56 \\
$\text{ReSC}_{L}^{\dag}$~\cite{resc} & DN53 & 77.63 & 80.45 & 72.30 & 63.59 & 68.36 & 56.81 & 63.12 & 67.30 & 67.20 & 36 \\
Iter-Shrinking~\cite{iterativeshrinking} & RN101 & - & 74.27 & 68.10 & - & 71.05 & 58.25 & - & - & 70.05 & - \\
$\text{LBYL}^{\dag}$~\cite{lbyl} & DN53 & 79.67 & 82.91 & 74.15 & 68.64 & 73.38 & 59.49 & 62.70 & - & - & 30 \\
TransVG~\cite{transvg} & RN101 & 81.02 & 82.72 & 78.35 & 64.82 & 70.70 & 56.94 & \underline{67.02} & 68.67 & 67.73 & 62 \\
$\text{TRAR}^{\dag}$~\cite{trar} & DN53 & - & 81.40 & \underline{78.60} & - & 69.10 & 56.10 & - & 68.90 & 68.30 & - \\
\noalign{\smallskip}
\hline
\noalign{\smallskip}
SeqTR~(ours) & DN53 & \underline{81.23} & \underline{85.00} & 76.08 & \underline{68.82} & \underline{75.37} & \underline{58.78} & - & \underline{71.35} & \underline{71.58} & 50 \\
$\text{SeqTR}^{\dag}$~(ours) & DN53 & \textbf{83.72} & \textbf{86.51} & \textbf{81.24} & \textbf{71.45} & \textbf{76.26} & \textbf{64.88} & \textbf{71.50} & \textbf{74.86} & \textbf{74.21} & 50 \\
\noalign{\smallskip}
\hline
\end{tabular}
\end{center}
\end{table*}
\setlength{\tabcolsep}{1.4pt}

\subsection{Evaluation Metrics}
\label{sect:evaluation metrics}
For REC and phrase localization, we evaluate the performance using Precision@0.5. The prediction is deemed correct if its intersection over union (IoU) with ground-truth box is larger than 0.5. For RES, we use \emph{mIoU} as the evaluation metric. Precision at 0.5, 0.7, and 0.9 thresholds are also used for ablation.

\setlength{\tabcolsep}{4.0pt}
\begin{table*}[t]
\scriptsize
\begin{center}
\caption{Comparison with the state-of-the-art models on the test set of Flickr30K Entities~\cite{flickrentities} and ReferItGame~\cite{referitgame} datasets.}
\label{table:sotaonphraselocalization}
\begin{tabular}{lcccc}
\hline\noalign{\smallskip}
\multirow{2}{*}{Models} & \multirow{2}{*}{\makecell{Visual \\ Encoder}} & ReferItGame & Flickr30k & Time \\
& & test & test & (ms) \\
\noalign{\smallskip}
\hline
\noalign{\smallskip}
\multicolumn{4}{l}{\textbf{Two-stage}} \\
\noalign{\smallskip}
\hline
\noalign{\smallskip}
MAttNet~\cite{mattnet} & RN101 & 29.04 & - & 320 \\
SimilarityNet~\cite{similaritynet} & RN101 & 34.54 & 60.89 & 184 \\
DDPN~\cite{ddpn} & RN101 & 63.00 & 73.30 & - \\
\noalign{\smallskip}
\hline
\noalign{\smallskip}
\multicolumn{4}{l}{\textbf{One-stage}} \\
\noalign{\smallskip}
\hline
\noalign{\smallskip}
FAOA~\cite{faoa} & DN53 & 60.67 & 68.71 & \textbf{23} \\
ZSGNet~\cite{zsgnet} & RN50 & 58.63 & 63.39 & - \\
RCCF~\cite{rccf} & DLA34 & 63.79 & - & 25 \\
$\text{ReSC}_{L}$~\cite{resc} & DN53 & 64.60 & 69.28 & 36 \\
TransVG~\cite{transvg} & RN101 & 70.73 & 79.10 & 62 \\
RefTR~\cite{reftr} & RN101 & \textbf{71.42} & 78.66 & 40 \\
\noalign{\smallskip}
\hline
\noalign{\smallskip}
SeqTR~(ours) & DN53 & 69.66 & \textbf{81.23} & 50 \\
\noalign{\smallskip}
\hline
\end{tabular}
\end{center}
\end{table*}
\setlength{\tabcolsep}{1.4pt}

\setlength{\tabcolsep}{3pt}
\begin{table*}[t]
\scriptsize
\begin{center}
\caption{Comparison with the state-of-the-arts on the RES task. Model with $*$ is pre-trained on the large corpus of data.}
\label{table:sotaonres}
\begin{tabular}{lcccc|ccc|ccc}
\hline\noalign{\smallskip}
\multirow{2}{*}{Models} & \multirow{2}{*}{\makecell{Visual \\ Encoder}} & \multicolumn{3}{c}{RefCOCO} & \multicolumn{3}{c}{RefCOCO+} & \multicolumn{3}{c}{RefCOCOg} \\
& & val & testA & testB & val & testA & testB & val-g & val-u & test-u \\
\noalign{\smallskip}
\hline
\noalign{\smallskip}
MAttNet~\cite{mattnet} & RN101 & 56.51 & 62.37 & 51.70 & 46.67 & 52.39 & 40.08 & - & 47.64 & 48.61 \\
CMSA~\cite{cmsa} & RN101 & 58.32 & 60.61 & 55.09 & 43.76 & 47.60 & 37.89 & 39.98 & - & - \\
STEP~\cite{step} & RN101 & 60.04 & 63.46 & 57.97 & 48.19 & 52.33 & 40.41 & 46.40 & - & - \\
BRINet~\cite{brinet} & RN101 & 60.98 & 62.99 & 59.21 & 48.17 & 52.32 & 42.11 & 48.04 & - & - \\
CMPC~\cite{cmpc} & RN101 & 61.36 & 64.53 & 59.64 & 49.56 & 53.44 & 43.23 & 49.05 & - & - \\
LSCM~\cite{lscm} & RN101 & 61.47 & 64.99 & 59.55 & 49.34 & 53.12 & 43.50 & 48.05 & - & - \\
CMPC+~\cite{cmpc+} & RN101 & 62.47 & 65.08 & 60.82 & 50.25 & 54.04 & 43.47 & 49.89 & - & - \\
MCN~\cite{mcn} & DN53 & 62.44 & 64.20 & 59.71 & 50.62 & 54.99 & 44.69 & - & 49.22 & 49.40 \\
EFN~\cite{efn} & WRN101 & 62.76 & 65.69 & 59.67 & 51.50 & 55.24 & 43.01 & \textbf{51.93} & - & - \\
BUSNet~\cite{busnet} & RN101 & 63.27 & 66.41 & 61.39 & 51.76 & 56.87 & 44.13 & 50.56 & - & - \\
CGAN~\cite{cgan} & DN53 & 64.86 & 68.04 & 62.07 & 51.03 & 55.51 & 44.06 & 46.54 & 51.01 & 51.69 \\
LTS~\cite{lts} & DN53 & 65.43 & 67.76 & 63.08 & 54.21 & 58.32 & 48.02 & - & 54.40 & 54.25 \\
VLT~\cite{vlt} & DN56 & 65.65 & 68.29 & 62.73 & 55.50 & 59.20 & 49.36 & 49.76 & 52.99 & 56.65 \\
\noalign{\smallskip}
\hline
\noalign{\smallskip}
SeqTR~(ours) & DN53 & 67.26 & 69.79 & 64.12 & 54.14 & 58.93 & 48.19 & - & 55.67 & 55.64 \\
$\text{SeqTR}^{*}$~(ours) & DN53 & \textbf{71.70} & \textbf{73.31} & \textbf{69.82} & \textbf{63.04} & \textbf{66.73} & \textbf{58.97} & - & \textbf{64.69} & \textbf{65.74} \\
\noalign{\smallskip}
\hline
\end{tabular}
\end{center}
\end{table*}
\setlength{\tabcolsep}{1.4pt}

\subsection{Implementation Details}
\label{sect:implementation details}

We train SeqTR 60 epochs for REC and phrase localization, and 90 epochs for RES with batch size 128. The Adam~\cite{adam} optimizer with an initial learning rate 5e-4 is used, which decays the learning rate 10 times after 50 epochs and 75 epochs for the detection and segmentation grounding tasks, respectively. Following standard practices~\cite{transvg,mcn,lts,vlt}, image size is resized to 640 $\times$ 640, and the length of language expression is trimmed at 15 for RefCOCO/+ and 20 for RefCOCOg. For ablation, we train SeqTR 30 epochs unless otherwise stated. During pre-training, SeqTR is trained 15 epochs and fine-tuned another 5 epochs. The number of quantization bins is set to 1000. We use DarkNet-53~\cite{yolov3} as the visual encoder. More details are provided in the appendix.

\subsection{Comparisons with State-of-the-Arts}
\label{sect:comparisons with state-of-the-art methods}

In this section, we compare the proposed SeqTR with the state-of-the-art methods on five benchmark datasets, \emph{i.e.}, RefCOCO, RefCOCO+, RefCOCOg, ReferItGame, and Flickr30K Entities. Tab.~\ref{table:sotaonrec} and Tab.~\ref{table:sotaonres} show the performance on REC and RES tasks. Tab.~\ref{table:sotaonpretrain} reports the result of SeqTR pre-trained on the large corpus of data. The performance on ReferItGame and Flickr30K Entities datasets are given in Tab.~\ref{table:sotaonphraselocalization}.

The performance of SeqTR on REC and phrase localization tasks is illustrated in Tab.~\ref{table:sotaonrec} and Tab.~\ref{table:sotaonphraselocalization}. From Tab.~\ref{table:sotaonrec}, our model performs better than two-stage models,  especially MAttNet~\cite{mattnet} while being 6 times faster. We also surpass one-stage models that exploit prior and expert knowledge, with +2-7\% absolute improvement over LBYL~\cite{lbyl} and ReSC~\cite{resc}. Despite we predict discrete coordinate tokens in an auto-regressive manner, the inference speed\footnote{Tested on GTX 1080 Ti GPU, batch size is 1.} of SeqTR is only 50ms, which is real-time and comparable with one-stage models. For transformer-based models, SeqTR surpasses TransVG~\cite{transvg} and TRAR~\cite{trar} with up to 6.27\% absolute performance improvement. Our SeqTR achieves new state-of-the-art performance with a simple architecture and loss function on the RefCOCO~\cite{refcoco/+}, RefCOCO+~\cite{refcoco/+}, and RefCOCOg~\cite{refcocoggoogle,refcocogumd} datasets. On the ReferItGame and Flickr30K Entities datasets which mostly contain short noun phrases, the performance boosts to 69.66 and 81.23 with a large margin over previous one-stage methods~\cite{faoa,zsgnet,rccf,resc} and is comparable with current state-of-the-art methods~\cite{transvg,reftr}. \par
SeqTR can be seamlessly extended to RES without any network architecture modifications since we reformulate the task as a point prediction problem. As shown in Tab.~\ref{table:sotaonres}, we outperform various models with sophisticated cross-modal alignment and reasoning mechanisms~\cite{lts,cgan,efn,mcn,cmsa,cmpc,cmpc+}. SeqTR is on par with current state-of-the-art VLT~\cite{vlt} which selectively aggregates responses from the diversified queries, whereas we directly produce the corresponding segmentation mask and establish one-to-one correspondence. When initialized with the pre-trained parameters using the large corpus of data, the performance boosts up to 10.78\% absolute improvement, proving that a simple yet universal approach for visual grounding is indeed feasible. \par
From Tab.~\ref{table:sotaonpretrain}, when pre-trained on the large corpus of text-image pairs, SeqTR is more data-efficient than the current state-of-the-art ~\cite{mdetr}. Our transformer architecture only contains 7.9M parameters which is twice as few as MDETR~\cite{mdetr}, while the performance is superior especially on the RefCOCOg dataset with up to 2.48\% improvement.

\subsection{Ablation Studies}
\label{sect:ablationstudies}

To give a comprehensive understanding of SeqTR, we discuss ablative studies on the validation set of the RefCOCO~\cite{refcoco/+}, RefCOCO+~\cite{refcoco/+}, and RefCOCOg~\cite{refcocogumd} datasets in this section.

\noindent{\bf Construction of language feature.} Language feature in Sec.~\ref{sect:architecture} can be constructed by either \emph{max}/\emph{mean} pooling of word features or directly using the final hidden state of bi-GRU. As shown in the upper part of Tab.~\ref{table:ablation-on-rec}, \emph{max} pooling performs best, and is the default construction throughout this paper.

\setlength{\tabcolsep}{1.2pt}
\begin{table*}[t]
\scriptsize
\begin{center}
\caption{Comparison with pre-trained models on RefCOCO~\cite{refcoco/+}, RefCOCO+~\cite{refcoco/+}, and RefCOCOg~\cite{refcocogumd} datasets. We only count the parameters of transformer architecture.}
\label{table:sotaonpretrain}
\begin{tabular}{lcc|c|ccc|ccc|cc}
\hline\noalign{\smallskip}
\multirow{2}{*}{Models} & \multirow{2}{*}{\makecell{Visual \\ Encoder}} & Params & \multirow{2}{*}{\makecell{Pre-train \\ images}} & \multicolumn{3}{c}{RefCOCO} & \multicolumn{3}{c}{RefCOCO+} & \multicolumn{2}{c}{RefCOCOg} \\
& & (M) & & val & testA & testB & val & testA & testB & val-u & test-u \\
\noalign{\smallskip}
\hline
\noalign{\smallskip}
ViLBERT~\cite{vilbert} & RN101 & - & 3.3M & - & - & - & 72.34 & 78.52 & 62.61 & - & - \\
$\text{VL-BERT}_{L}$~\cite{vlbert} & RN101 & - & 3.3M & - & - & - & 72.59 & 78.57 & 62.30 & - & - \\
$\text{UNITER}_{L}$~\cite{uniter} & RN101 & - & 4.6M & 81.41 & 87.04 & 74.17 & 75.90 & 81.45 & 66.70 & 74.86 & 75.77 \\
$\text{VILLA}_{L}$~\cite{villa} & RN101 & - & 4.6M & 82.39 & 87.48 & 74.84 & 76.17 & 81.54 & 66.84 & 76.18 & 76.71 \\
$\text{ERNIE-ViL}_{L}$~\cite{ernie} & RN101 & - & 4.3M & - & - & - & 75.95 & 82.07 & 66.88 & - & - \\
MDETR~\cite{mdetr} & RN101 & 17.36 & 200K & 86.75 & 89.58 & 81.41 & \textbf{79.52} & 84.09 & 70.62 & 81.64 & 80.89 \\
RefTR~\cite{reftr} & RN101 & 17.86 & \textbf{100K} & 85.65 & 88.73 & 81.16 & 77.55 & 82.26 & 68.99 & 79.25 & 80.01 \\
\noalign{\smallskip}
\hline
\noalign{\smallskip}
SeqTR~(ours) & DN53 & \textbf{7.90} & 174K & \textbf{87.00} & \textbf{90.15} & \textbf{83.59} & 78.69 & \textbf{84.51} & \textbf{71.87} & \textbf{82.69} & \textbf{83.37} \\
\noalign{\smallskip}
\hline
\end{tabular}
\end{center}
\end{table*}
\setlength{\tabcolsep}{1.2pt}

\setlength{\tabcolsep}{4pt}
\begin{table*}[t]
\scriptsize
\begin{center}
\caption{Ablation experiments on the construction of language feature and token weight. The first token is the $\left[\text{TASK}\right]$ token, while subsequent tokens are discrete coordinate tokens, \emph{i.e.}, $(x_1, y_1, x_2, y_2)$.}
\label{table:ablation-on-rec}
\begin{tabular}{c|ccccc|c|c|c}
\hline\noalign{\smallskip}
     \multirow{2}{*}{Language feature}& \multicolumn{5}{c}{Token weight} & RefCOCO & RefCOCO+ & RefCOCOg \\
     & 1st & 2nd & 3rd & 4th & 5th & val & val & val-u \\
\noalign{\smallskip}
\hline
\noalign{\smallskip}
     mean pooling & \multirow{3}{*}{1} & \multirow{3}{*}{1} & \multirow{3}{*}{1} & \multirow{3}{*}{1} & \multirow{3}{*}{1} & 79.73 & 67.12 & 68.97 \\
     max pooling &  &  &  &  &  & {\bf 80.07} & {\bf 68.31} & {\bf 69.95} \\
     final hidden state &  &  &  &  &  & 79.85 & 67.46 & 69.93 \\
\noalign{\smallskip}
\hline
\noalign{\smallskip}
     \multirow{7}{*}{max pooling} & 1 & 1 & 1 & 1 & 1 & 80.07 & 68.31 & 69.95 \\
     & 1.5 & 1 & 1 & 1 & 1 & 80.10 & {\bf 68.63} & {\bf 70.05} \\
     & 2 & 1 & 1 & 1 & 1 & {\bf 80.19} & 68.33 & 70.01 \\
     & 3 & 1 & 1 & 1 & 1 & 80.08 & 67.81 & 69.45 \\
     & 1 & 2 & 2 & 1 & 1 & 79.70 & 67.22 & 69.51 \\
     & 2 & 2 & 2 & 1 & 1 & 80.16 & 67.83 & 69.45 \\
\noalign{\smallskip}\hline
\end{tabular}
\end{center}
\end{table*}
\setlength{\tabcolsep}{1.4pt}

\begin{figure*}[ht!]
\centering
\includegraphics[height=5.78cm, width=12.2cm]{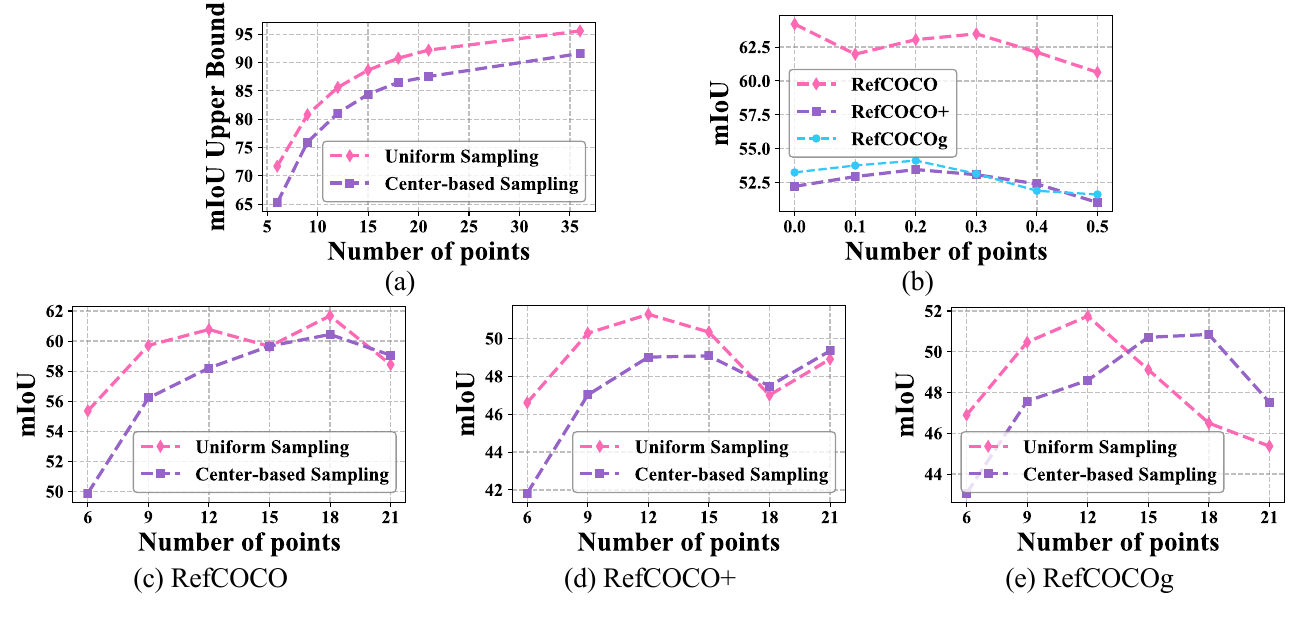}
\caption{Ablative experiments on RES task. (a) The upper bound is averaged over validation sets (the fluctuation is within 0.2). (b) Shuffling percentage refers to the fraction of shuffled sequences within a batch, uniform sampling strategy is used. (c-e) depict the impact of sampling strategies and the number of sampled points.}
\label{fig:abl-on-res}
\end{figure*}

\noindent{\bf Token weight.} If previously predicted points are inaccurate, model can not recover from the wrong predictions since the inference is sequential. Hence we increase a few former token weights to penalize more on the first several predicted discrete coordinate tokens. As shown in the lower part of Tab.~\ref{table:ablation-on-rec}, increasing the weight of first token is better than increasing the latter tokens, and setting the 1st token weight to 1.5 and subsequent tokens to 1 gives the best performance. We set $w_i=1, \forall i$ for RES task.

\noindent{\bf Sampling scheme.} We verify the upper bound as the mIoU of the assembled mask from the sampled points and original ground-truth. From Fig.~\ref{fig:abl-on-res} (a), we can see that the mIoU approaches nearly 100 when the number of sampled points increases, \emph{i.e.}, 95.57 for uniform sampling, and 91.58 for center-based sampling. Therefore, though the upper bound is limited theoretically, in practice, the research effort might be better spent on improving the real-world performance. In terms of sampling strategies, from Fig.~\ref{fig:abl-on-res} (a) and Fig.~\ref{fig:abl-on-res} (c-e), uniform sampling is consistently better than center-based sampling in terms of both the upper bound and the performance, which preserves more details of the mask illustrated in Fig.~\ref{fig:sampling}. The number of sampled points controls the trade-off between the inference speed and performance, from Fig.~\ref{fig:abl-on-res} (c-e), we can see that 18 and 12 points are the best for RefCOCO and RefCOCO+/RefCOCOg datasets.

\noindent{\bf Shuffling percentage.} We train SeqTR 60 epochs instead of 30 as we empirically found that point shuffling takes a longer time to converge, since the ground-truth is different for each coordinate token at each forward pass. Fig.~\ref{fig:abl-on-res} (b) shows that no shuffle and 0.2 are best for RefCOCO and RefCOCO+/RefCOCOg. As the number of shuffled sequences increases, the performance drops slightly, and we observe that SeqTR is under-fitting since the mIoU during training is lower than the one without shuffling.

\noindent{\bf Multi-task training.} Previous multi-task visual grounding approaches require REC to help RES locate the referent. In contrast, \emph{SeqTR is capable to locate the referent at pixel level without the aid from REC}. We train SeqTR 60 epochs and test whether multi-task supervision can bring further improvement. For the input sequence construction of multi-task grounding, please see the supplementary material. From Tab.~\ref{table:ablation-on-multi-task}, we can see that multi-task supervision even slightly degenerates the performance compared to the single-task variant. Though the inconsistency error significantly decreases, the location ability of RES measured by Prec@0.5, 0.7, and 0.9 stays the same, suggesting that the sampled points are independent between the sequence of the bounding box and binary mask.

\subsection{Qualitative Results}
\label{sect:qualitative results}

\begin{figure*}[ht]
\centering
\includegraphics[height=5.51cm, width=\textwidth]{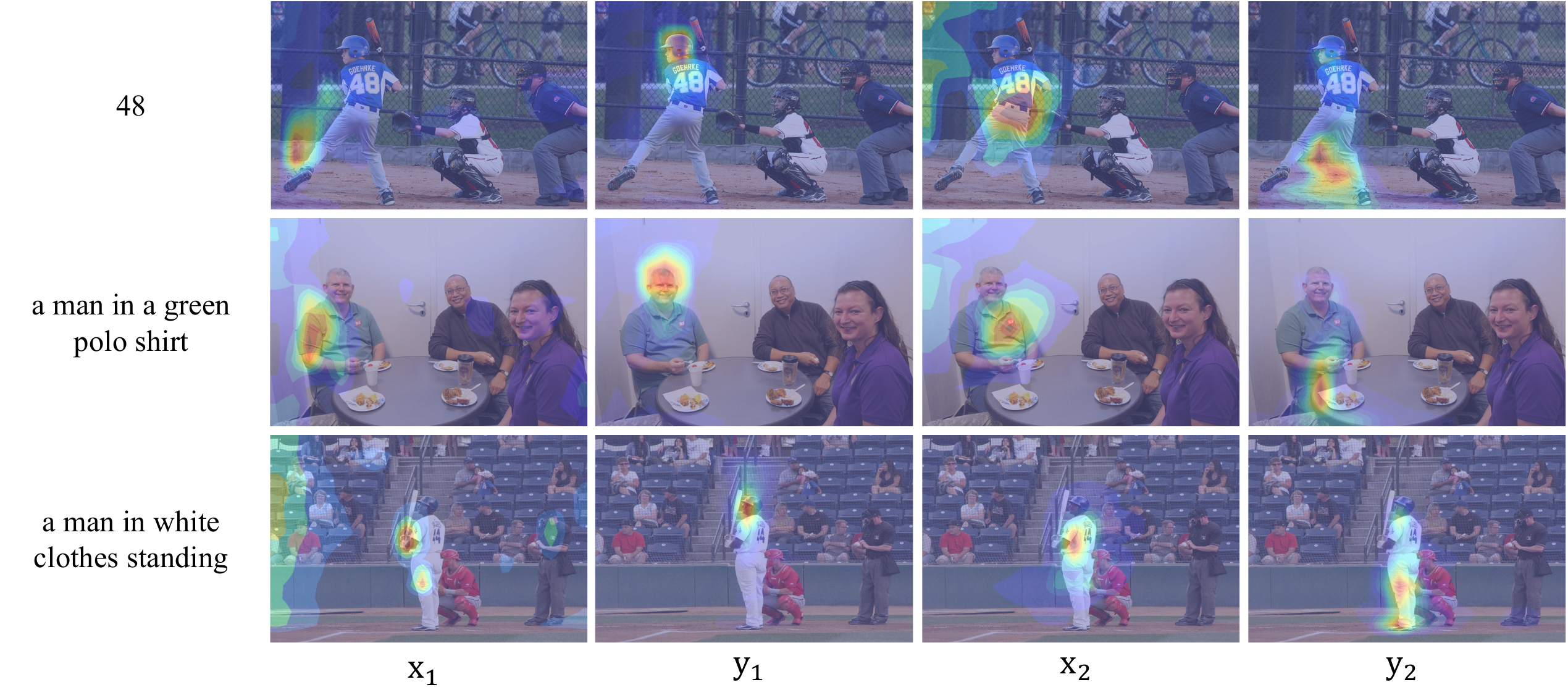}
\caption{Visualization of normalized cross attention map in transformer decoder. From left to right column, we generate $(x_1, y_1, x_2, y_2)$ in sequential order.}
\label{fig:attention}
\end{figure*}

\begin{figure*}[ht]
\centering
\includegraphics[height=4.85cm, width=11.77cm]{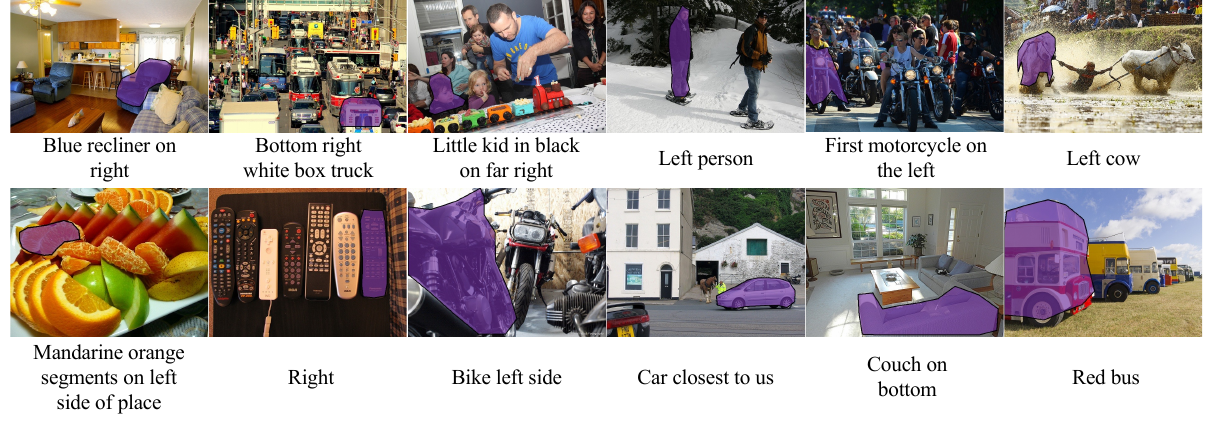}
\caption{Example mask predictions by SeqTR on the validation set of RefCOCO dataset, best viewed in color.}
\label{fig:mask}
\end{figure*}

We visualize the cross attention map averaged over decoder layers and attention heads in Fig.~\ref{fig:attention}. At each prediction step, SeqTR generates a coordinate token given previous output tokens. Under this setting, a clear pattern emerges, \emph{i.e.}, attends to the left side of the referent when predicting $x_1$, the top side of the referent when predicting $y_1$, and so on. This axial attention is sensitive to the boundary of the referent, thus can more precisely ground the referred object. The predicted masks are visualized in Fig.~\ref{fig:mask}. SeqTR can well comprehends attributive words and absolute or relative spatial relations, and the predicted mask aligns with the irregular outlines of the referred object such as ``\emph{left cow}''. More qualitative results are given in the appendix.

\setlength{\tabcolsep}{3pt}
\begin{table*}[t]
\scriptsize
\begin{center}
\caption{Ablation study of multi-task training. IE is the inconsistency error~\cite{mcn} to measure the prediction conflict between REC and RES, $\downarrow$ denotes the lower is better.}
\label{table:ablation-on-multi-task}
\begin{tabular}{cccccccc}
\hline\noalign{\smallskip}
     \multirow{2}{*}{Dataset} & \multirow{2}{*}{Multi-task training} & REC & \multicolumn{3}{c}{RES} & \multirow{2}{*}{mIoU} & \multirow{2}{*}{IE$\downarrow$} \\
     & & Prec@0.5 & Prec@0.5 & Prec@0.7 & Prec@0.9 & \\
\noalign{\smallskip}
\hline
\noalign{\smallskip}
     \multirow{2}{*}{RefCOCO} & \XSolidBrush & {\bf 80.38} & {\bf 78.03} & {\bf 63.35} & {\bf 9.75} & {\bf 64.20} & 13.93 \\
     & \CheckmarkBold & 79.65 & 77.24 & 60.29 & 7.23 & 62.93 & {\bf 5.86} \\
\noalign{\smallskip}
\hline
\noalign{\smallskip}
    \multirow{2}{*}{RefCOCO+} & \XSolidBrush & 67.98 & 65.11 & 48.27 & 5.19 & 52.22 & 22.22 \\
    & \CheckmarkBold & {\bf 68.79} & {\bf 66.67} & {\bf 51.02} & {\bf 5.46} & \bf{53.65} & {\bf 4.85} \\
\noalign{\smallskip}
\hline
\noalign{\smallskip}
    \multirow{2}{*}{RefCOCOg} & \XSolidBrush & 69.63 & {\bf 65.20} & {\bf 46.23} & {\bf 5.31} & {\bf 53.25} & 22.65 \\
    & \CheckmarkBold & {\bf 70.29} & {\bf 65.20} & 46.05 & 5.15 & {\bf 53.25} & {\bf 8.25} \\
\noalign{\smallskip}
\hline
\end{tabular}
\end{center}
\end{table*}
\setlength{\tabcolsep}{1.4pt}

\section{Conclusions}
\label{sect:conclusions}

In this paper we reformulate visual grounding tasks as a point prediction problem and present an innovative and general network termed SeqTR. Based on the standard transformer encoder-decoder architecture and \emph{cross-entropy} loss, SeqTR unifies different visual grounding tasks under the same point prediction paradigm without any modifications. Experimental results demonstrate that SeqTR can well ground language query onto the corresponding region, suggesting that a simple yet universal approach for visual grounding is indeed feasible. 

\subsubsection{Acknowledgements.}

This work was supported by the National Science Fund for Distinguished Young Scholars (No. 62025603), the National Natural Science Foundation of China (No. U21B2037, No. 62176222, No. 62176223, No. 62176226, No. 62072386, No. 62072387, No. 62072389, and No. 62002305), Guangdong Basic and Applied Basic Research Foundation (No. 2019B1515120049), and the Natural Science Foundation of Fujian Province of China (No. 2021J01002).



\clearpage

%
%
\bibliographystyle{splncs04}
\bibliography{SeqTR}

\clearpage

\appendix
\renewcommand{\thesection}{\Alph{section}}
\section{Appendix}

\subsection{More implementation details}

Exponential moving average (EMA) with a decay rate of 0.999 is used to accelerate training convergence following ~\cite{mdetr}. In contrast to previous methods~\cite{faoa,resc,transvg}, in which random color distortion, affine transformation, and horizontal flipping are used to augment the image, we do not perform any data augmentation except large scale jittering (LSJ)~\cite{copypaste} following ~\cite{pix2seq}, with jittering strength of 0.3 to 1.4. EMA and LSJ are disabled during pre-training and ablation studies. Label Smoothing with a smoothing factor of 0.1 is used to regularize the predictor. It takes nearly a day to train for 60 epochs on a single V100 GPU without mixed precision training.

\subsection{Sequence construction for Multi-task grounding}

\begin{figure*}[ht]
\centering
\includegraphics[height=3.06cm, width=7.5cm]{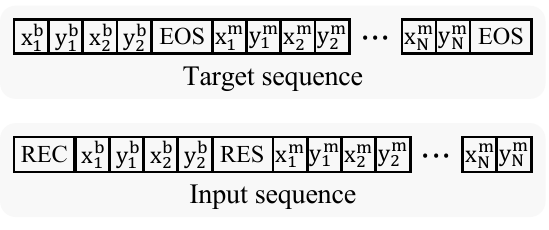}
\caption{Sequence construction from the bounding box and binary mask for multi-task visual grounding. $\left[\text{REC}\right]$ and $\left[\text{RES}\right]$ are the $\left[\text{TASK}\right]$ tokens randomly initialized with different parameters. Coordinates with superscript $b$ are for the bounding box and $m$ for the binary mask.}
\label{fig:multi-task}
\vspace{-2em}
\end{figure*}

We construct the input and target sequence for the transformer decoder as shown in Fig.~\ref{fig:multi-task} when perform multi-task visual grounding. The construction is similar compared to the single-task variant except that there are two distinct $\left[\text{TASK}\right]$ tokens, one for the grounding task at bounding box level, \emph{i.e.}, REC or phrase localization, and the other for the grounding task at pixel level. As discussed in the paper, multi-task training does not improve the performance, hence, we report the results of the single-task trained performance.

\subsection{Nucleus sampling for RES}

During inference, each predicted discrete coordinate token is the \emph{argmax}-ed index over the normalized probabilities, here we study the impact of the stochastic Nucleus Sampling~\cite{pix2seq,nucleus} strategy widely used in natural language generation community, which reduces duplication and increases the diversity in the predicted sequence. As shown in Tab.~\ref{table:ablation-on-nucleus}, nucleus sampling does not improve the quality of generated sequence representing the predicted binary mask and introduces an additional hyper-parameter $p$, hence, we use argmax in the paper.

\setlength{\tabcolsep}{4pt}
\begin{table*}[ht]
\scriptsize
\begin{center}
\caption{The effect of $p$ in nucleus sampling, which samples from a truncated ranked list of discrete coordinate tokens. Setting $p$ to 0 equals to \emph{argmax} selection.}
\label{table:ablation-on-nucleus}
\begin{tabular}{cccc}
\hline\noalign{\smallskip}
     \multirow{2}{*}{top-$\emph{p}$} & RefCOCO & RefCOCO+ & RefCOCOg \\
     & val & val & val-u \\
\noalign{\smallskip}
\hline
\noalign{\smallskip}
    0.0 & \textbf{67.26} & 54.14 & \textbf{55.67} \\
    0.1 & 66.76 & 54.66 & 55.54 \\
    0.2 & 66.72 & \textbf{54.78} & 55.49 \\
    0.3 & 66.68 & 54.71 & 55.46 \\
    0.4 & 66.50 & 54.60 & 55.37 \\
    0.5 & 66.38 & 54.34 & 55.08 \\
    0.6 & 66.15 & 54.04 & 54.79 \\
\noalign{\smallskip}
\hline
\vspace{-3em}
\end{tabular}
\end{center}
\end{table*}
\setlength{\tabcolsep}{1.4pt}

\subsection{More qualitative results}

\begin{figure*}[t]
\vspace{-12em}
\centering
\includegraphics[height=14.11cm, width=\textwidth]{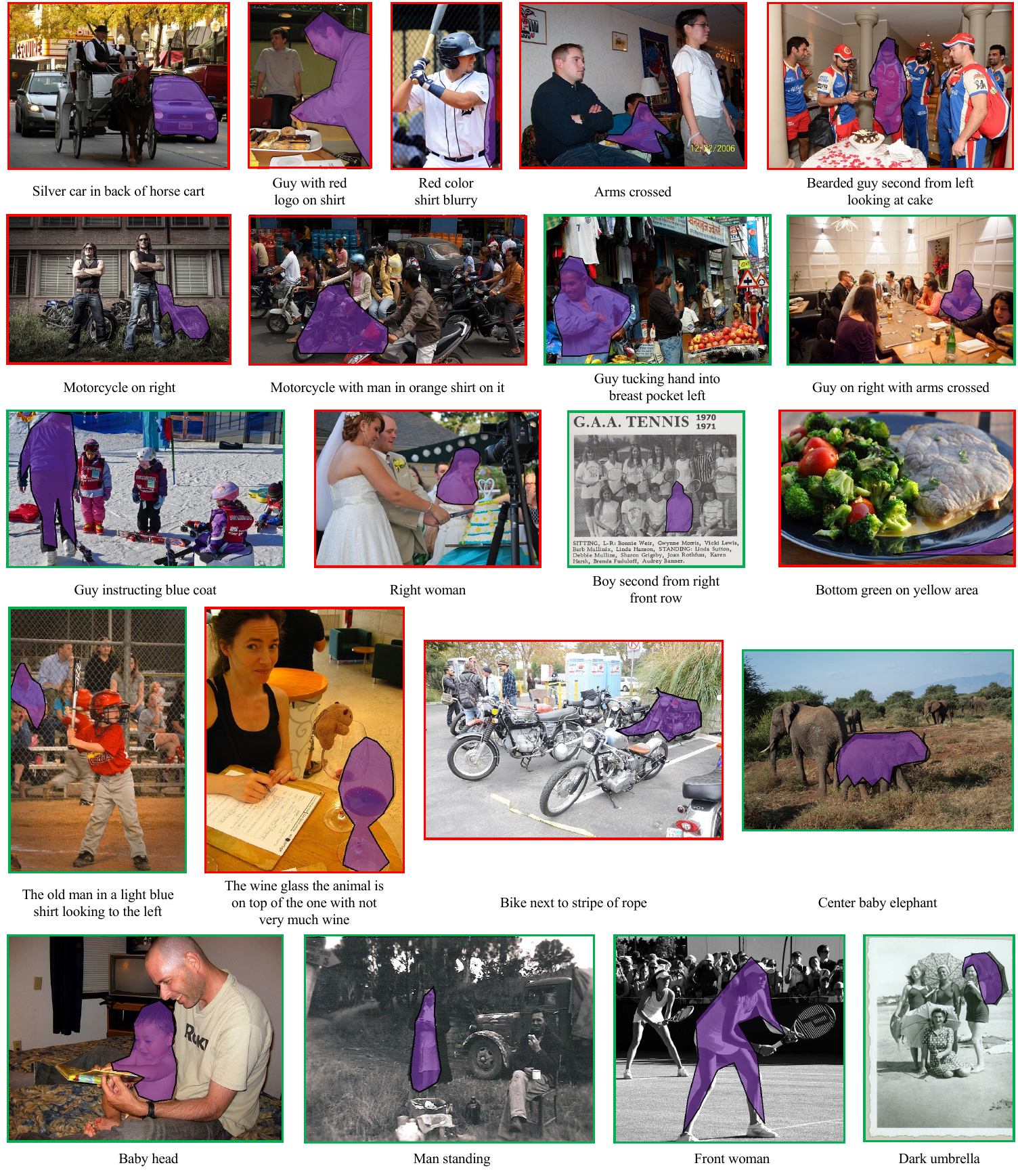}
\caption{Visualizations of the predicted masks. Ground-truth binary masks can be inferred from the language query.}
\label{fig:more-vis}
\end{figure*}

As shown in Fig.~\ref{fig:more-vis}, the wrong predictions (marked with red box) can be mainly divided into two groups, \emph{i.e.}, the prediction either shifts to the objects of the same category with the referent but is not referenced in the query, or only aligns with the largest segment of the referent. The first case can be addressed using a better multi-modal fusion module to suppress the salient objects. However, to demonstrate the efficacy of our overall network, we do not resort to such a potentially complex fusion module. When the ground-truth binary mask contains multiple segments, \emph{i.e.}, occluded by other objects, we only find the contour of the largest segment and sample points atop of it, while discard other segments, this results in SeqTR only grounding the query onto the largest segment of the mask instead of our model's incapability of segmentation.

\end{document}